%% file: neurips_2025.tex
\documentclass{article}

\usepackage[preprint]{neurips_2025}

\usepackage[utf8]{inputenc} % allow utf-8 input
\usepackage[T1]{fontenc}    % use 8-bit T1 fonts
\usepackage{hyperref}       % hyperlinks
\usepackage{url}            % simple URL typesetting
\usepackage{booktabs}       % professional-quality tables
\usepackage{amsfonts}       % blackboard math symbols\
\usepackage{nicefrac}       % compact symbols for 1/2, etc.
\usepackage{microtype}      % microtypography
\usepackage{xcolor}         % colors
\usepackage{amssymb}
\usepackage{amsmath}
\usepackage{mathtools}
\usepackage{amsthm}
\usepackage{bm}
\usepackage{graphicx}
\usepackage{subcaption}
\usepackage[ruled,vlined]{algorithm2e}

\DontPrintSemicolon
\SetKwInput{KwIn}{Input}
\SetKwInput{KwOut}{Output}
\renewcommand{\KwIn}{\textbf{Input:} }

\SetKwProg{Fn}{Function}{:}{end} % define Function ... end
\SetKwComment{Comment}{// }{}   % inline comments with //
\SetKwRepeat{Repeat}{repeat}{until} % repeat ... until

\title{A Minimalist Method for Fine-tuning Text-to-Image Diffusion Models}

\author{%
  Yanting Miao\\
  Department of Computer Science \\
  University of Waterloo, Vector Institute\\
  \texttt{y43miao@uwaterloo.ca} \\
  \And
  William Loh \\
  Department of Computer Science \\
  University of Waterloo, Vector Institute \\
  \texttt{wmloh@uwaterloo.ca} \\
  \And
  Pascal Poupart \\
  Department of Computer Science \\
  University of Waterloo, Vector Institute \\
  \texttt{ppoupart@uwaterloo.ca} \\
    \And
  Suraj Kothawade \\
  Google Research \\
  \texttt{skothawade@google.com} 
}

\begin{document}

\maketitle

\begin{abstract}
Recent work uses reinforcement learning (RL) to fine-tune text-to-image diffusion models, improving text–image alignment and sample quality. However, existing approaches introduce unnecessary complexity: they cache the full sampling trajectory, depend on differentiable reward models or large preference datasets, or require specialized guidance techniques. Motivated by the ``golden noise'' hypothesis—that certain initial noise samples can consistently yield superior alignment—we introduce Noise PPO, a minimalist RL algorithm that leaves the pre-trained diffusion model entirely frozen and learns a prompt-conditioned initial noise generator. Our approach requires no trajectory storage, reward backpropagation, or complex guidance tricks. Extensive experiments show that optimizing the initial noise distribution consistently improves alignment and sample quality over the original model, with the most significant gains at low inference steps. As the number of inference steps increases, the benefit of noise optimization diminishes but remains present. These findings clarify the scope and limitations of the golden noise hypothesis and reinforce the practical value of minimalist RL fine-tuning for diffusion models.

\end{abstract}

\input{sections/intro}
\input{sections/background}
\input{sections/method}

\input{sections/implementation}

\input{sections/experiment}
\input{sections/conclusion}

\bibliographystyle{plainnat}
\bibliography{ref}

\input{sections/appendix}

\end{document}

%% file: sections/intro.tex
% Introduction section template for NeurIPS paper
% File: intro.tex

%========================================================
% Introduction
%========================================================

\section{Introduction}
\label{sec:intro}
Deep generative models, particularly diffusion models \citep{sohl2015deep, ho2020denoising, song2020score, song2020denoising} and flow-matching methods \citep{liu2022flow, lipman2022flow}, have achieved remarkable progress in image generation \citep{rombach2022high, saharia2022photorealistic}, video synthesis \citep{ho2022video, blattmann2023align}, and molecular design \citep{hoogeboom2022equivariant, xu2022geodiff}. Despite their success, diffusion models often fall short in aligning with downstream requirements such as text–image correspondence \citep{Kirstain2023PickaPicAO}, human preference \citep{xu2023imagereward, wu2023human}, and aesthetic quality \citep{schuhmann2022laion}.

In the large language model (LLM) community, reinforcement learning (RL) fine-tuning—such as PPO \citep{schulman2017proximal}, DPO \citep{rafailov2023direct}, and GRPO \citep{shao2024deepseekmath}—has become a standard approach for aligning model outputs with reward signals. Inspired by these advances, recent work has begun to apply RL to diffusion models. However, existing RL-based fine-tuning methods for diffusion models often introduce significant complexity: they require storing full sampling trajectories \citep{black2023training}, rely on differentiable reward models \citep{clark2023directly} or large-scale preference datasets \citep{wallace2024diffusion}, or depend on specialized guidance techniques \citep{li2024reneg}. These requirements not only increase memory and computational costs, but also limit the generality and reproducibility of such methods.
Amidst these developments, the ``golden noise'' hypothesis \citep{qi2024not, zhou2024golden} has emerged, suggesting that certain initial noise samples can consistently yield superior text–image alignment in diffusion models. This raises a fundamental question: 
\begin{quote}
    \textit{Can optimizing only the initial noise distribution provide the benefits of RL fine-tuning, and under what conditions does this hold?}
\end{quote}

To investigate this hypothesis, we propose Noise PPO, a minimalist RL-based algorithm that learns a prompt-conditioned initial noise generator for text-to-image diffusion models, while keeping the pre-trained diffusion model entirely frozen. By focusing solely on the initial noise, Noise PPO sidesteps the need for trajectory storage, reward backpropagation, and complex guidance tricks, offering a simple and practical RL-based fine-tuning method.

Through extensive experiments, we find that optimizing the initial noise distribution consistently improves text–image alignment and sample quality over the baseline model. Notably, zero-initialization of the policy yields strong and robust improvements across most metrics and settings, making it a reliable default choice. However, non-zero-initialization can lead to even higher gains in aesthetic quality, and this advantage persists even as the number of inference steps increases. These findings clarify the scope and limitations of the golden noise hypothesis, and provide new insights into the role of initial noise in diffusion models and the practical boundaries of RL-based fine-tuning.

\begin{figure}
    \centering
    \includegraphics[width=\linewidth]{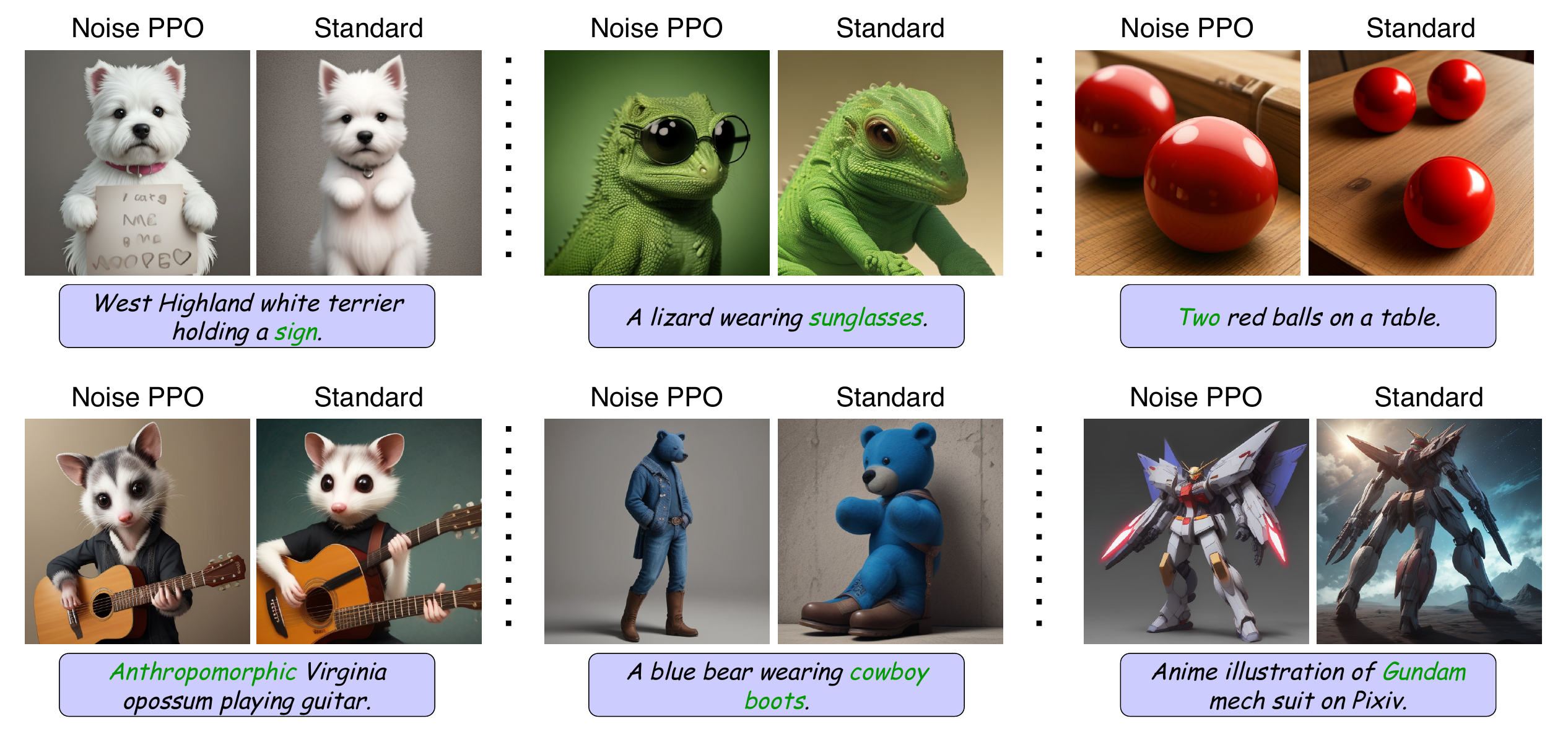}
    \caption{We propose Noise PPO, a simple reinforcement learning algorithm for fine-tuning diffusion models by learning a prompt-conditioned policy to generate initial noise. Notably, this approach requires no modification to the pre-trained diffusion model. ``Standard'' denotes images generated using standard Gaussian noise, while ``Noise PPO'' uses noise sampled from the learned policy. For each prompt, images are generated with the same random seed for a fair comparison}
    \label{fig:comparison}
\end{figure}

%% file: sections/background.tex
\section{Background}

\paragraph{Diffusion model and flow-matching.} Diffusion models \citep{sohl2015deep, song2020score, ho2020denoising} and flow matching \citep{liu2022flow, lipman2022flow} are powerful classes of deep generative models. These approaches define a continuous process, often governed by a stochastic differential equation (SDE) or an ordinary differential equation (ODE), that transforms samples from a simple initial distribution $p_0$, typically, $p_0(\xv) = \Nc(\xv; \mathbf{0}, \Iv)$, to samples approximating a complex target data distribution $p_{\text{data}}$. This process evolves over $t \in [0, 1]$, starting from $\xv_0 \sim p_0$ and resulting in $\xv_1$ whose distribution $p_1$ aims to match $p_{\text{data}}$.

A common formulation specifies the conditional probability path $p_t(\xv_t | \xv_1)$ as a Gaussian distribution $\Nc(\xv_t; \alpha_t \xv_1, \beta_t^2 \Iv)$, where $\alpha_t$ and $\beta_t$ are noise scheduling parameters that satisfy the boundary conditions $\alpha_0 = 0$, $\alpha_1 = 1$, $\beta_0 = 1$, and $\beta_1 = 0$. The marginal probability path can be defined by $p_t(\xv_t) = \Eb_{\xv_1 \sim p_{\text{data}}}[p_t(\xv_t \mid \xv_1)].$

Given $\xv_t$, the diffusion model aims to train a neural network to estimate the scaled negative score function of the marginal probability path, expressed as $\epsilonv_{\psiv}(\xv_t, t) \approx -\beta_t \nabla_{\xv} \log p_{t}(\xv_t)$. Accordingly, the training objective involves learning the network $\epsilonv_{\psiv}(\xv_t, t)$ by minimizing the denoising score matching loss:
\begin{align}
    \min_{\psiv} \Eb \left[ \lVert \epsilonv_{\psiv}(\alpha_t \xv_1 + \beta_t \epsilonv_t, t) - \epsilonv_t \rVert^2\right],
\end{align}
where the expectation is taken with $t \sim \text{Unif}[0, 1], \xv_1 \sim p_{\text{data}}(\xv_1), \epsilonv_t \sim \Nc(\epsilonv_t; \mathbf{0}, \Iv)$. Unless otherwise specified, we use $\Eb[\cdot]$ to denote the expectation over all random variables throughout this paper. Text-to-image diffusion models use the same denoising objective as their unconditional counterparts, but are augmented with a text encoder $f_{\psiv}$ and cross-attention layers to incorporate the prompt. We denote the resulting text-to-image pipeline by $\Psiv(\xv_0, \yv; \psiv)$, where $\yv$ is the input text prompt.

\paragraph{Reinforcement Learning.} Reinforcement learning \citep{sutton1998reinforcement} formalizes sequential decision‑making as a Markov decision process (MDP), $\Mc = (\Sc, \Ac, \Pc, \Rc, H)$, where $\Sc$ is the state space, $\Ac$ the action space, $\Pc: \Sc \times \Ac \to \Sc$ the transition function, $\Rc: \Sc \times \Ac \to \Rb$ the reward function, and $H$ the horizon. A policy $\pi$ may be stochastic, mapping state $\sv \in \Sc$ to a distribution over actions, or deterministic, mapping state $\sv$ directly to an action $\av$. The RL objecive is to maximize the expected return $\Eb[\sum_{t=0}^H \Rc(\sv_t, \av_t)]$ when following the policy in the MDP. Given a state, the expected cumulative reward can be captured by a state-value function, $V^\pi(\sv_t) \coloneqq \Eb[\sum_{h=t}^H \Rc(\sv_h, \av_h)]$. In this work, we consider the one-step RL $(H = 1)$, where there is no temporal accumulation and the state-value function simplifies to $V^\pi(\sv) = \Eb_{\av \sim \pi(\av \mid \sv)}[\Rc(\sv, \av)].$

A more detailed discussion of related work can be found in Appendix~\ref{sec:related_work}.

%% file: sections/method.tex
\section{Method}

We introduce Noise PPO, a minimalist RL framework designed to test the golden noise hypothesis in diffusion models. Unlike prior RL-based approaches \citep{black2023training, fan2023dpok} that treat the entire diffusion process as a multi-step Markov decision process (MDP), Noise PPO reduces the problem to a one-step RL task by optimizing only the initial noise distribution, while keeping the pre-trained diffusion model completely frozen. This design not only simplifies implementation and training, but also allows us to directly probe the impact of initial noise on generation quality.
\paragraph{RL formulation.}
We formalize the task as a one-step RL problem:
\begin{align*}
    &\sv = \yv \quad \av = \xv_0 \quad  \pi(\av \mid \sv) = p_0(\xv_0) = \Nc(\xv_0; \mathbf{0}, \Iv) \\
    &\Rc(\sv, \av) = \Rc(\yv, \xv_0) = \sum_i^M w_i R_i(\yv, \xv_1), \text{ where } \xv_1 = \Psiv(\xv_0, \yv; \psiv),
\end{align*}
where $\yv$ is the text prompt, $\xv_0$ is the initial noise, and $\xv_1$ is the generated image. Each reward model $R_i(\yv, \xv_1)$ evaluates the sample with respect to the prompt, and $w_i$ are scaling weights. The diffusion model acts as part of the environment, mapping initial noise and prompt to the final image.

\begin{figure}
    \centering
    \includegraphics[width=1.0\linewidth]{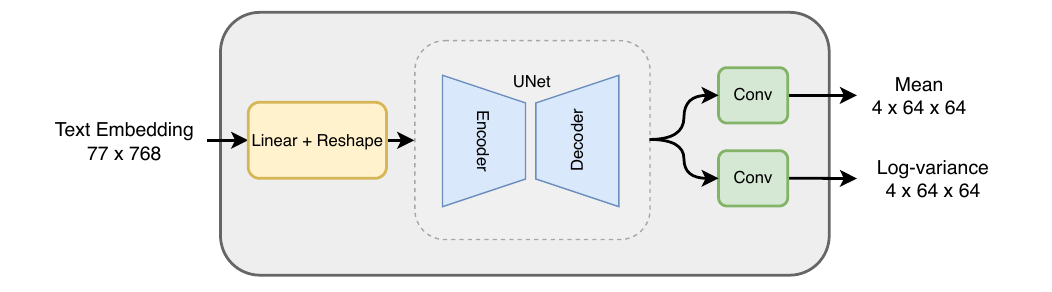}
    \caption{Architecture of the policy network: a text embedding obtained from the pre-trained diffusion model’s text encoder is first projected via a linear layer into a low-resolution feature map. This map is processed by a UNet and then fed into two separate $1 \times 1$ convolution heads, one producing the mean and the other producing the log-variance of the initial noise distribution.}
    \label{fig:policy_net_arch}
\end{figure}

\paragraph{Policy.}
In standard diffusion, the initial noise is sampled from a prompt-agnostic standard normal distribution. In contrast, we learn a prompt-conditioned noise policy $\pi_{\thetav}(\xv_0 \mid \yv)$, modeled as a Gaussian:
$$\pi_{\thetav}(\xv_0 \mid \yv) = \Nc(\xv_0; \mu_{\thetav}(\yv), \Sigmav_{\thetav}(\yv)).$$
Given the deterministic nature of the ODE solver, once $\xv_0$ is chosen, the final image and reward are fully determined. Figure~\ref{fig:policy_net_arch} illustrates the policy network, which is a compact UNet with two $1 \times 1$ convolution heads for mean and log-variance.

\paragraph{Reward models.}
The reward function combines text–image alignment (using HPSv2 \citep{wu2023human} and PickScore \citep{Kirstain2023PickaPicAO}) and aesthetic quality (using the LAION aesthetic predictor \citep{schuhmann2022laion}). This composite reward encourages generations that are both faithful to the prompt and visually appealing. Unlike some prior work \citep{qi2024not}, our method does not require denoising-inversion to measure the cosine similarity between the original noise and its inversion, which would unnecessarily increase training cost.

\paragraph{Training.}
We train the prompt-conditioned noise policy using proximal policy optimization (PPO) \citep{schulman2017proximal}. The PPO objective for a single sample $(\xv_0, \yv)$ is:
\begin{align}
    &\Jc_{\text{PPO}}(\thetav; \xv_0, \yv, \thetav_{\text{old}}) = \min \bigg\{\frac{\pi_{\thetav}(\xv_0 \mid \yv)}{\piold(\xv_0 \mid \yv)}A^{\piold}(\yv, \xv_0), \nonumber \\
    &\text{clip} \left(\frac{\pi_{\thetav}(\xv_0 \mid \yv)}{\piold(\xv_0 \mid \yv)}, 1 - \epsilon, 1 + \epsilon \right)A^{\piold}(\yv, \xv_0) \bigg\}, \label{eq:ppo_obj} \\
    & \qquad A^{\piold}(\yv, \xv_0) = \Rc(\yv, \xv_0) - V^{\piold}(\yv),
\end{align}
where $\epsilon$ controls the trust region and $V^{\piold}$ is the old-policy value estimation. To prevent the learned policy from drifting too far from the standard Gaussian prior $\Nc(\xv_0; \mathbf{0}, \Iv)$, we add KL divergence $\Db_{\text{KL}}(\pi_{\thetav}(\xv_0 \mid \yv) \| \Nc(\xv_0; \mathbf{0}, \Iv))$ as a penalty term. Further, inspired by the max entropy RL, we also include an entropy bonus $\Hc(\pi_{\thetav}(\xv_0 \mid \yv))$. Combining these terms, the overall policy loss over a prompt dataset is 
\begin{align}\label{eq:rl_obj}
   \min_{\thetav} \Eb_{\substack{\yv \sim \Dc \\ \xv_0 \sim \piold(\xv_0 \mid \yv)}}\left[-\Jc_{\text{PPO}}(\thetav; \xv_0, \yv, \thetav_{\text{old}}) + \gamma_1 \Db_{\text{KL}}(\pi_{\thetav}(\cdot \mid \yv) \| \Nc(\cdot; \mathbf{0}, \Iv)) - \gamma_2 \Hc(\pi_{\thetav}(\cdot \mid \yv)) \right],
\end{align}
where $\gamma_1$ and $\gamma_2$ control the KL and entropy terms. We use a separate simple MLP $V_{\phiv}(\yv)$ for value estimation. Although GRPO does not require training an additional value network, selecting an appropriate group size presents certain challenges. If the group size is too small, it may introduce bias in the estimation of the state-value function. Conversely, if the group size is too large, it can significantly increase training time due to the more demanding sampling process. We leave GRPO fine-tuning for future work. Algorithm~\ref{alg:noise_ppo} (in Appendix~\ref{sec:imp_details}) provides a summary of our Noise PPO method.

%% file: sections/implementation.tex
\section{Implementation}\label{sec:implementation}
We describe our implementation details in this section. The core ideas of Noise PPO are model-agnostic and can be applied to a variety of diffusion architectures. Additional implementation details, i.e. hyperparameters, are provided in Appendix~\ref{sec:imp_details}.

\paragraph{Diffusion pipeline.}
Since Noise PPO is an on-policy algorithm, sampling efficiency is crucial. We adopt the Latent Consistency Model (LCM) \citep{luo2023lcm} as our main diffusion pipeline, which is distilled from the pre-trained Stable Diffusion v1.5 model \citep{rombach2022high}. LCM enables fast inference with a small number of steps; by default, we set the number of inference steps to 4 during training.
\paragraph{Text encoder.}
We use the publicly available text encoder from Stable Diffusion v1.5 to encode prompts for both the policy and the value function. For the policy $\pi_{\thetav}(\xv_0 \mid \yv)$, we use the last hidden state of shape $\mathbb{R}^{77 \times 768}$ as input. For the value function $V_{\phiv}(\yv)$, we use the pooled hidden state (i.e., the $\mathsf{[CLS]}$ token) of shape $\mathbb{R}^{768}$.

\paragraph{Policy network.}
The policy network adopts a UNet-style architecture to predict the mean and log-variance of the initial noise distribution, conditioned on the text prompt. The last hidden state from the frozen text encoder is linearly projected and reshaped into a spatial 2D grid to match the UNet input format. The network follows an encoder–bottleneck–decoder structure with skip connections. Each encoder and decoder block contains two convolutional layers, each followed by GELU activation and group normalization. The bottleneck consists of two additional convolutional layers. Upsampling in the decoder is performed using transposed convolutions. The final output heads are two separate $1 \times 1$ convolutions that produce the mean and log-variance, both initialized to zero so that the initial policy matches the standard Gaussian. By default, the network has four encoder and decoder blocks, with a maximum channel width of 1024 at the bottleneck. The total number of trainable parameters is approximately 187 million.

\paragraph{Value network.}
The value network is a simple feedforward neural network that estimates the state-value function from the pooled text embedding. It consists of three linear layers with GELU activations, mapping the input embedding to a scalar value.

%% file: sections/experiment.tex
\section{Experiment}\label{sec:experiment}
In this section, we conduct a comprehensive empirical study to address the following key questions:
\begin{itemize}
    \item \textbf{Effectiveness.} Can Noise PPO effectively improve the performance of diffusion models, and how does it compare to existing fine-tuning methods?
    \item \textbf{Initialization.} Is initializing the policy to produce standard Gaussian noise necessary for successful training, or can alternative initializations also yield good results?
    \item \textbf{Generalization.} Although the policy is trained with a fixed number of inference steps, can it generalize to unseen inference steps at test time? How does its performance compare to the standard Gaussian baseline under varying sampling conditions?
\end{itemize}
We first describe our experimental setup, including datasets, evaluation metrics, and baselines, and then present our main results and analysis. We further evaluate the applicability of Noise PPO on alternative diffusion pipelines (e.g., SDXL-Turbo \citep{sauer2024adversarial}), with additional quantitative and qualitative comparison results provided in Appendix~\ref{sec:more_results}.

\subsection{Experimental Setup}
\paragraph{Training dataset.}
Unlike most existing fine-tuning methods for diffusion models, our approach does not require paired (prompt, image) training data. Instead, we train solely on the OpenPrompt dataset \citep{openprompts}, which contains high-quality prompts for text-to-image generation. Since our goal is to optimize the initial noise distribution rather than the full generative process, we use only a small, curated subset of 10,000 prompts, filtering out those that are too short or too long.

\paragraph{Evaluation metrics.}
We evaluate model performance using the same reward functions employed during RL training: the LAION aesthetic score \citep{schuhmann2022laion}, HPSv2 \citep{wu2023human}, and PickScore \citep{Kirstain2023PickaPicAO}. Our main results are reported on two widely used benchmarks:
\begin{itemize}
    \item \textbf{PartiPrompts benchmark} \citep{yu2022scaling}: 1,632 prompts spanning diverse categories, including animals, indoor scenes, art, and more.
    \item \textbf{HPSv2 benchmark} \citep{wu2023human}: Four subdomains (animation, concept art, painting, photography), each with 800 prompts, enabling comprehensive evaluation of both text-image alignment and aesthetic quality.
\end{itemize}

\paragraph{Comparison methods.}
We compare Noise PPO against the following representative approaches:
\begin{itemize}
\item \textbf{Pre-trained models:} SD v1.5 \citep{rombach2022high} and LCM \citep{luo2023lcm}.
\item \textbf{RL-based fine-tuning:} Direct Preference Optimization for Diffusion Models (DDPO) \citep{black2023training} and Diffusion-DPO \citep{wallace2024diffusion}.
\item \textbf{Prompt and embedding refinement:} DNP \citep{desai2024improving} (auxiliary negative prompts), ReNeg \citep{li2024reneg} (unconditional embedding optimization), and TextCraftor \citep{li2024textcraftor} (text encoder fine-tuning).
\end{itemize}

\begin{table}
    \centering
    \caption{Comparison results on PartiPrompts benchmark, including Aesthetic Score, PickScore, HPSv2, and Total Score. Noise PPO improves all metrics relative to its base model (LCM). $\uparrow$ indicates that Noise PPO outperforms the base model (LCM).\\
    $^{\dag}$\scriptsize{Results for previous methods are reported from \citet{li2024textcraftor}. The ReNeg checkpoint is re-evaluated using the model provided by \citet{li2024reneg}.}}
    \begin{tabular}{l|l l l l}
        \hline
        Method & Aesthetic & PickScore & HPSv2 & Total \\ \hline
        SD v1.5 & 5.26 & 18.83 & 27.07 & 51.16 \\
        LCM & 5.86 & 21.89 & 27.54 & 55.29 \\ \hline
        Diffusion-DPO & 5.26 & 19.48 & 26.62 & 51.36 \\ 
        DDPO & 5.26 & 18.70 & 26.76 & 50.72 \\
        DNP & 5.21 & 19.81 & 25.83 & 50.85 \\
        ReNeg & 5.57 & 21.36 & \textbf{28.13} & 55.05 \\
        TextCraftor & 5.88 & 19.16 & 28.05 & 53.09 \\ \hline
        Noise PPO (non-zero init) & \textbf{5.96} $\uparrow$ & 21.99 $\uparrow$ & 27.55 $\uparrow$ & 55.50 $\uparrow$ \\
        Noise PPO (zero init) & 5.88 $\uparrow$ & \textbf{22.00} $\uparrow$ & 27.70 $\uparrow$ & \textbf{55.58} $\uparrow$ \\ \hline
    \end{tabular}
    \label{tab:parti_results}
\end{table}

\begin{table}
    \centering
    \caption{Comparison results on the HPSv2 benchmark. For fairness, we exclude methods that are fine-tuned solely on the HPSv2 score. Noise PPO achieves leading results and improves all metrics relative to its base model (LCM). $\uparrow$ indicates that Noise PPO outperforms the baseline model (LCM).}
    \begin{tabular}{l|l l l l l}
        \hline
        Method & Anime & Concept Art & Painting & Photo & Average \\ \hline
        SD v1.5 & 27.21 & 26.83 & 26.86 & 27.75 & 27.27 \\
        LCM & 28.03 & 27.26 & 27.55 & 27.60 & 27.61 \\ \hline
        Diffusion-DPO & 27.60 & 26.42 & 26.36 & 26.32 & 26.67 \\ 
        DDPO & 20.45 & 20.53 & 20.12 & 20.33 & 20.36 \\ 
        DNP & 26.02 & 25.08 & 24.89 & 25.49 & 25.37 \\ \hline
        Noise PPO (non-zero init) & 28.08 $\uparrow$ & \textbf{27.60} $\uparrow$ & 27.72 $\uparrow$ & 27.79 $\uparrow$ & 27.80 $\uparrow$ \\
        Noise PPO (zero init) & \textbf{28.17} $\uparrow$ & 27.58 $\uparrow$ & \textbf{27.73} $\uparrow$ & \textbf{27.88} $\uparrow$ & \textbf{27.84} $\uparrow$ \\
        \hline
    \end{tabular}
    \label{tab:hpsv2_results}
\end{table}

\begin{figure}
    \centering
    \includegraphics[width=\linewidth]{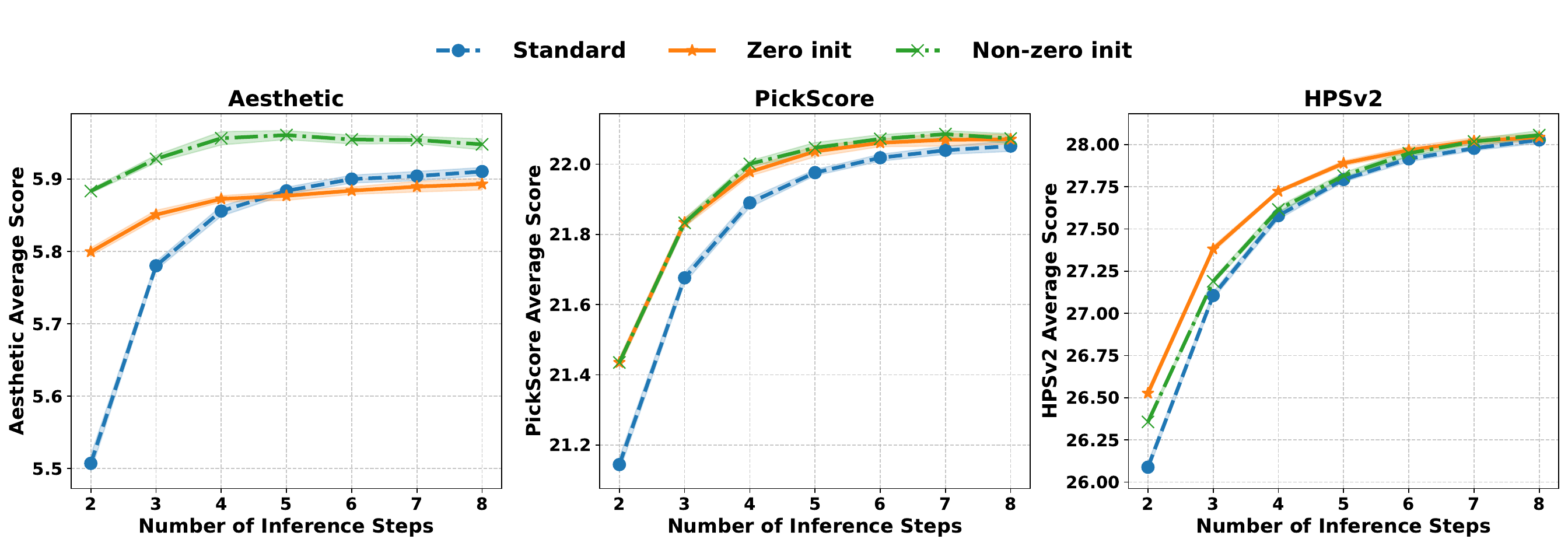}
    \caption{Performance on the PartiPrompts benchmark for Aesthetic Score, PickScore, and HPSv2 as a function of the number of inference steps. Noise PPO (zero and non-zero initialization) outperforms the standard Gaussian baseline, with the performance gap narrowing as the number of inference steps increases. Results are measured over 10 different seeds, with the standard deviation (std) indicated by the translucent shaded region.}
    \label{fig:diff_infer_parti}
\end{figure}

\begin{figure}
    \centering
    \includegraphics[width=0.7\linewidth]{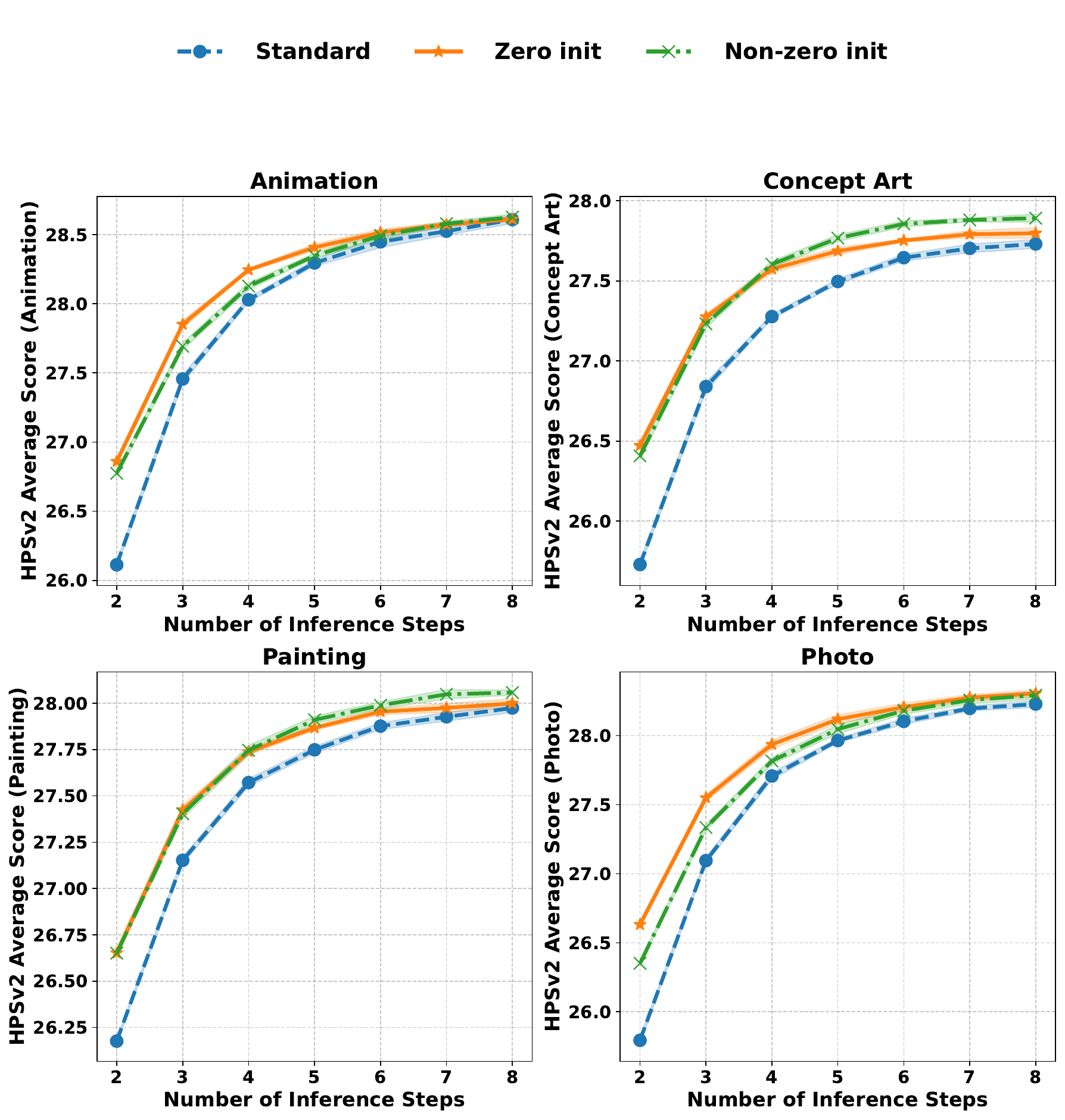}
    \caption{HPSv2 reward as a function of inference steps for four subdomains: Animation, Concept Art, Painting, and Photo. Noise PPO (zero and non-zero initialization) outperforms the standard Gaussian baseline, with the performance gap narrowing as the number of inference steps increases. Results are measured over 10 different seeds, with the standard deviation (std) indicated by the translucent shaded region.}
    \label{fig:diff_infer_hpsv2}
\end{figure}

% \begin{figure}[htbp]
%     \centering
%     \begin{subfigure}{0.45\textwidth}
%         \includegraphics[width=\linewidth]{figures/inference_step_curve_lcm_ppo_hpsv2_Animation.pdf}
%         \caption{Animation}
%     \end{subfigure}
%     \begin{subfigure}{0.45\textwidth}
%         \includegraphics[width=\linewidth]{figures/inference_step_curve_lcm_ppo_hpsv2_Concept_Art.pdf}
%         \caption{Concept art}
%     \end{subfigure}
%     \vspace{0.5em}
%     \begin{subfigure}{0.45\textwidth}
%         \includegraphics[width=\linewidth]{figures/inference_step_curve_lcm_ppo_hpsv2_Painting.pdf}
%         \caption{Painting}
%     \end{subfigure}
%     \begin{subfigure}{0.45\textwidth}
%         \includegraphics[width=\linewidth]{figures/inference_step_curve_lcm_ppo_hpsv2_Photo.pdf}
%         \caption{Photo}
%     \end{subfigure}
%     \caption{HPSv2 reward as a function of inference steps for four subdomains: (a) Animation, (b) Concept Art, (c) Painting, and (d) Photo. Noise PPO (zero and non-zero initialization) consistently outperforms the standard Gaussian baseline, with the performance gap narrowing as the number of inference steps increases.}
%     \label{fig:diff_infer_hpsv2}
% \end{figure}

\begin{figure}
    \centering
    \includegraphics[width=1.0\linewidth]{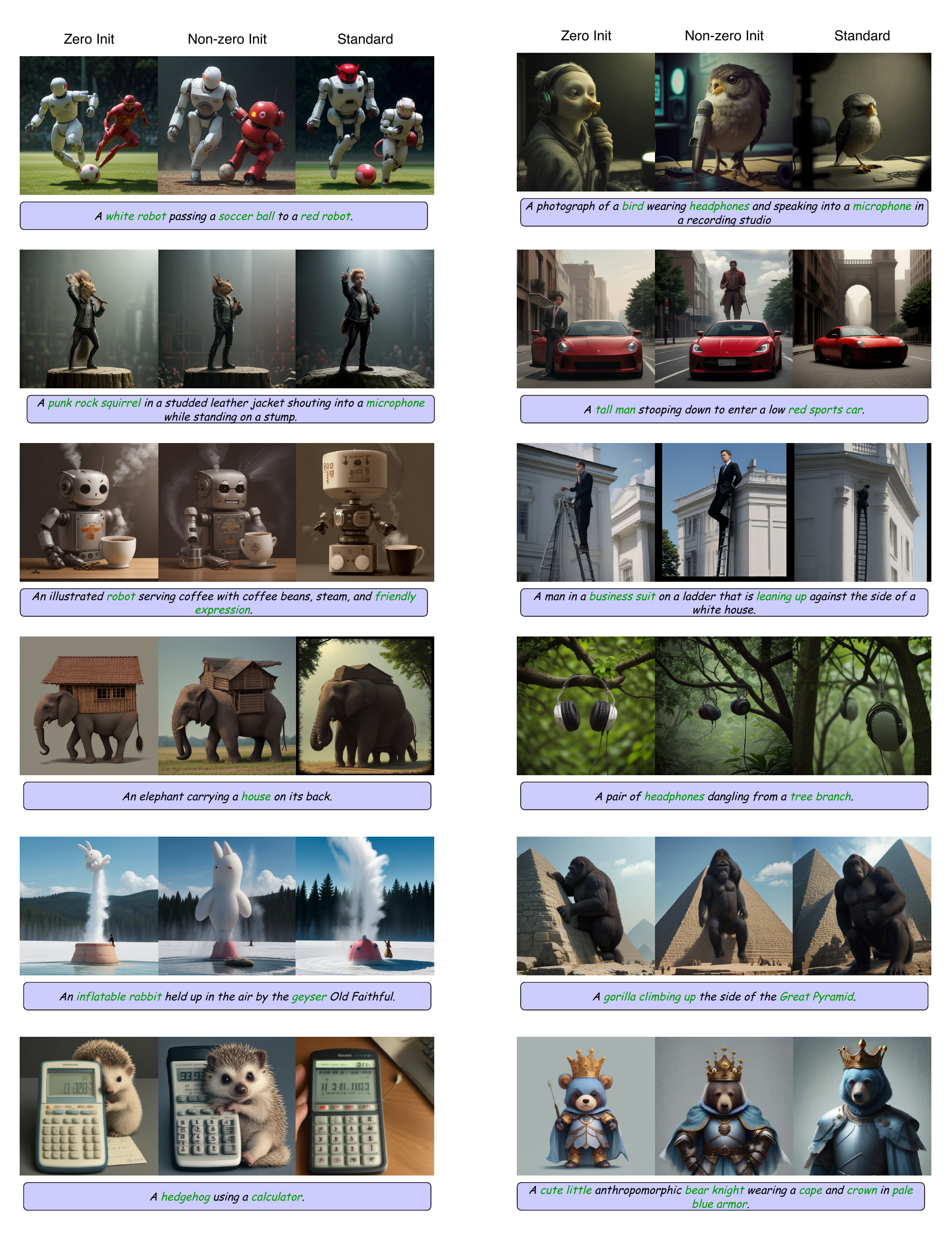}
    \caption{Qualitative visualizations. ``Zero Init'' denotes experiments where the final output layer is initialized to zero, while ``Non-zero Init'' refers to random initialization of the output layer. ``Standard'' corresponds to images generated from standard Gaussian noise. We observe that images generated with standard Gaussian noise consistently exhibit the poorest visual quality. In contrast, applying Noise PPO—regardless of the initialization scheme—improves both text-image alignment and aesthetic quality over the standard Gaussian baseline. For instance, under the top-left robot prompt, Noise PPO generates a correctly aligned red robot without the one-legged artifact observed in images produced with standard Gaussian noise, regardless of initialization.}
    \label{fig:zero_vs_nonzero_vs_standard}
\end{figure}

\subsection{Effectiveness and Initialization of Noise PPO}
\paragraph{Quantitative comparison.}
We first evaluate the overall effectiveness of Noise PPO and examine the impact of different policy initialization strategies. We report zero-shot evaluation results for all three reward metrics on the PartiPrompts benchmark. As shown in Table~\ref{tab:parti_results}, both zero-initialized and non-zero-initialized versions of Noise PPO consistently outperform the baseline LCM model across all evaluation metrics, including Aesthetic Score, PickScore, HPSv2, and Total Score. These results demonstrate that optimizing the initial noise distribution via RL yields tangible improvements over the original diffusion model, regardless of the initialization scheme.

Notably, the performance gains are robust to the choice of initialization, indicating that Noise PPO is both effective and stable in practice. We observe similar trends on the HPSv2 benchmark (see Table~\ref{tab:hpsv2_results}), where both initialization strategies lead to consistent improvements over the baseline across all subdomains. This further confirms the general effectiveness of Noise PPO in enhancing diffusion model performance and supports the existence of a golden noise generator—an optimized initial noise distribution that consistently yields better results.

\citet{li2024textcraftor} point out that DDPO has limited capability to generalize to unseen prompts due to its reliance on early stopping, making it difficult to balance overfitting and performance improvement during training. Furthermore, Diffusion-DPO still depends on log-density estimation, and computing the ELBO is not an optimal solution for the DPO objective. In contrast, training Noise PPO is straightforward: we know that the optimal initial distribution should be close to the standard Gaussian, so massive amounts of training data are not required. Additionally, since we model the distribution as Gaussian, the log-density needed for policy gradient computation is easy to evaluate. This simplicity, combined with the empirical effectiveness of Noise PPO, highlights its practical value as a tool for discovering and leveraging the golden noise generator in diffusion models.

\paragraph{Qualitative comparison.}
We illustrate the generative quality of Noise PPO in Figures~\ref{fig:comparison} and~\ref{fig:zero_vs_nonzero_vs_standard}, where images are generated with the same random seed for fair comparison. These results show that sampling quality is substantially improved compared to LCM with standard Gaussian noise. In the first row of Figure~\ref{fig:comparison}, we observe failure cases caused by standard Gaussian initial noise, while Noise PPO significantly enhances text-image alignment. The second row demonstrates that Noise PPO also improves aesthetic quality by removing artifacts (e.g., the ``three-legged Gundam'' example). 

Across visualizations, both zero-initialized and non-zero-initialized Noise PPO produce images that are more faithful to the prompt and contain fewer artifacts compared to the standard Gaussian baseline. Notably, zero-initialized Noise PPO often demonstrates superior text-image alignment. For example, in the top right panel of Figure~\ref{fig:zero_vs_nonzero_vs_standard}, the image generated with standard Gaussian noise depicts only the bird, omitting both the headphones and the microphone. The non-zero-initialized Noise PPO captures the bird and microphone but fails to include the headphones. In contrast, only the zero-initialized Noise PPO successfully renders all elements described in the text prompt. A similar pattern appears in the fifth row of Figure~\ref{fig:zero_vs_nonzero_vs_standard}: while the non-zero-initialized Noise PPO generates a gorilla, it does not capture the “climbing up” behavior, which is accurately depicted by the zero-initialized Noise PPO. These results suggest that the optimal golden noise generator should be close to, but not identical to, the standard Gaussian distribution. Overall, these observations further support the conclusion that Noise PPO effectively discovers a golden noise generator, leading to consistently improved generative outcomes.

\subsection{Generalization Across Inference Steps}
To evaluate the generalization ability of Noise PPO, we test the learned noise policy across a range of inference steps, including those not seen during training. Figure~\ref{fig:diff_infer_parti} presents the results on the PartiPrompts benchmark, showing the performance for Aesthetic Score, PickScore, and HPSv2 as a function of the number of inference steps. In all three metrics, both zero-initialized and non-zero-initialized Noise PPO policies consistently outperform the standard Gaussian baseline. The performance gap is most significant at lower inference steps and gradually narrows as the number of steps increases.

However, we observe an interesting exception for the Aesthetic Score: the non-zero-initialized Noise PPO maintains a substantial advantage over the baseline across all inference steps, with its aesthetic reward consistently higher than both the standard Gaussian and zero-initialized policies. This persistent improvement suggests that, for the aesthetic reward, there may exist a form of ``golden noise generator'' that remains effective even as the number of inference steps increases.

We observe similar trends on the HPSv2 benchmark, as shown in Figure~\ref{fig:diff_infer_hpsv2}, which reports the HPSv2 reward for four subdomains: Animation, Concept Art, Painting, and Photo. In each subdomain, Noise PPO achieves higher scores than the baseline at every inference step, with the most pronounced improvements at low step counts. As the number of inference steps increases, the advantage of Noise PPO generally diminishes; however, the non-zero-initialized policy often retains a lead, especially for aesthetic-related metrics such as the concept art and painting subdomains.

This phenomenon can be attributed to the nature of the diffusion process. When the number of inference steps is small, the initial noise has a stronger influence on the final generated image, making the optimization of the initial noise distribution particularly effective. As the number of steps increases, the iterative denoising process gradually reduces the effect of the initial noise, causing the performance of different initializations to become closer. However, the persistent advantage of the non-zero-initialized policy for the aesthetic reward indicates that, at least for certain objectives, a golden noise can be found that generalizes well across sampling regimes. This highlights the nuanced applicability of the golden noise hypothesis, which may depend on the specific reward being optimized.

%% file: sections/conclusion.tex
\section{Conclusion}
In this work, we revisited the golden noise hypothesis and introduced Noise PPO, a minimalist RL framework for optimizing the initial noise distribution in diffusion models. Our experiments show that Noise PPO consistently enhances text-to-image alignment and sample quality over the baseline, with the most significant gains at low inference steps. Notably, the non-zero-initialized policy often discovers a golden noise generator that generalizes well across sampling regimes.
These results clarify the impact and limitations of initial noise optimization, suggesting it is a simple yet effective way to boost diffusion model performance, especially under certain reward and sampling settings.

%% file: sections/appendix.tex
\appendix

\section{Implementation Details} \label{sec:imp_details}
\paragraph{Training}
By default, both the policy and value networks are optimized using the AdamW optimizer \citep{loshchilov2017decoupled} for 10,000 gradient steps, with a weight decay of $1 \times 10^{-5}$ and momentum parameters $(0.9, 0.999)$. Training is conducted with a batch size of 16, gradient accumulation over 4 steps, and a fixed learning rate of $1 \times 10^{-4}$. A constant learning rate schedule is employed throughout. The reward weights are $(0.2, 0.4, 0.4)$ for aesthetic scores, PickScore, and HPSv2 scores, respectively. All models are trained on a single NVIDIA A100 GPU (80GB), requiring approximately 20 hours to complete. Additional hyperparameters are provided in Table~\ref{tab:training_hyperparameters}.

\paragraph{Pseudo-code.} See Algorithm ~\ref{alg:noise_ppo}.

\begin{algorithm}[H] \label{alg:noise_ppo}
\caption{Noise PPO}
\KwIn{Prompt dataset $\Dc$, pre-trained text-to-image diffusion model $\Psiv(\cdot, \cdot; \psiv)$, PPO epoch $K$, hyperparameters $\gamma_1$ and $\gamma_2$} \;
    Initialize policy parameters $\thetav$ and value function parameters $\phiv$ \;
    \Repeat{converged}{
        $\thetav_{\text{old}} \gets \thetav$ \;
        Sample a set of prompts $\yv \sim \Dc$ \;
        Sample a set of initial noise $\xv_0 \sim \piold (\xv_0 \mid \yv)$ \;
        Obtain reward $r \gets \Rc(\yv, \xv_0)$ \;
        $\Dc_{\text{RL}} \gets \{\xv_0, \yv, r, \log \piold(\xv_0 \mid \yv)\}$ \;
        \For{$k = 1, \dots, K$}{
            Sample a mini-batch $(\xv_0, \yv, r, \log \piold(\xv_0 \mid \yv)) \sim \Dc_{\text{RL}}$ \;
            Take a gradient descent step on
            $$\nabla_{\phiv} \left(V_{\phiv}(\yv) - r\right)^2$$ \;
            Take a gradient descent step on 
            $$\nabla_{\thetav} -\Jc_{\text{PPO}}(\thetav; \xv_0, \yv, \thetav_{\text{old}}) + \gamma_1 \Db_{\text{KL}}(\pi_{\thetav}(\cdot \mid \yv) \| \Nc(\cdot; \mathbf{0}, \Iv)) -\gamma_2 \Hc(\pi_{\thetav}(\cdot \mid \yv))$$
        }
    }
\end{algorithm}

\section{Related Work}\label{sec:related_work}

\paragraph{Generative models.}
Denoising diffusion models \citep{sohl2015deep, ho2020denoising, song2020score, song2020denoising}, flow-matching models \citep{liu2022flow, lipman2022flow}, and consistency models \citep{song2023consistency} have become powerful generative frameworks for tasks ranging from high-quality image synthesis \citep{saharia2022photorealistic, rombach2022high} and video generation \citep{ho2022video, blattmann2023align} to drug design \citep{schneuing2024structure} and robot trajectory modelling \citep{janner2022planning, chi2023diffusion}. Diffusion models, in particular, learn to transform samples from a standard Gaussian into complex data distributions by minimizing the evidence lower bound (ELBO) of the log-likelihood. However, \citet{karras2024guiding} observe that this ELBO objective drives models to cover the entire training distribution, often at the expense of sample fidelity and prompt alignment. As a result, even state-of-the-art generative models can produce outputs that lack sharpness or fail to fully respect their conditioning inputs.

\paragraph{Controllable generation for diffusion models.} 
A common technique for aligning diffusion outputs with textual conditions is classifier-free guidance (CFG) \citep{ho2022classifier}, which steers samples toward higher conditional likelihoods by interpolating between conditional and unconditional models. DNP \citep{desai2024improving} replaces the unconditional “null” token with negative prompts generated by another vision–language model \citep{achiam2023gpt}, while ReNeg \citep{li2024reneg} learns a separate unconditional text embedding to guide the diffusion process. However, \citet{karras2024guiding, kynkaanniemi2024applying, zheng2023characteristic} show that CFG’s denoising trajectory does not correspond to a valid diffusion toward the true data distribution. To eliminate reliance on CFG, recent work has distilled new conditional models that inherit CFG’s alignment properties \citep{tang2025diffusion} or employed ``auto-guidance'', using a perturbed version of the same model to self-guide generation \citep{karras2022elucidating}.

\paragraph{Reinforcement learning fine-tuning.} 
Several recent studies have applied RL to fine-tune large pre-trained models. In the language domain, GPT-based systems \citep{achiam2023gpt} and DeepSeek \citep{guo2025deepseek} use PPO and GRPO to improve reasoning capabilities. In text-to-image generation, DDPO \citep{black2023training} and DPOK \citep{fan2023dpok} adapt PPO and DPO to enhance the aesthetic quality of diffusion models. These methods rely on the Gaussian transition $p_{\thetav}(\xv_{t + \Delta t} \mid \xv_t)$ in DDPM,  but cannot be applied to deterministic ODE samplers such as DDIM \citep{song2020denoising} or EDM \citep{karras2022elucidating}, where $\xv_{t + \Delta t}$ is a deterministic function of $\xv_t$, and the log-density is ill-defined. Diffusion-DPO \citep{wallace2024diffusion} extends preference-based fine-tuning to the SDXL model \citep{podell2023sdxl}, but requires a large, annotated human-preference dataset. DRaFT \citep{clark2023directly} backpropagates gradients through the entire sampling trajectory, incurring high computational and memory costs. 

\paragraph{Inversion process.} Whereas the diffusion sampling process maps noise to data, the inversion process reverses this mapping, reconstructing noise from data. \citet{hertz2022prompt} introduce DDIM inversion, which traverses the diffusion dynamics backward, from $\xv_1$ to $\xv_0$ via
\begin{align}
    \xv_{t - \Delta t} = \frac{\alpha_{t - \Delta t}}{\alpha_t} \xv_t - \left(\frac{\alpha_{t - \Delta t}}{\alpha_t} \beta_t - \beta_{t - \Delta t}\right) \epsilonv_{\psiv}(\xv_t, t).
\end{align}
We denote the full denoising–inversion mapping by $\Fc(\xv_0)$, i.e., $\xv_0 \xrightarrow{\text{denoise}} \xv_1 \xrightarrow{\text{invert}} \xv_0^\prime$. Empirical studies, \citep{qi2024not, zhou2024golden} indicate that a high cosine similarity between $\xv_0$ and $\Fc(\xv_0)$ correlates with improved text-image alignment.

\paragraph{Initial noise for diffusion models.}

Recent studies by \citet{qi2024not} and \citet{zhou2024golden} show that certain ``golden'' noise vectors outperform samples drawn uniformly from a standard Gaussian. \citep{qi2024not} hypothesis that a golden noise $\xv_0$ should have high cosine similarity with its denoising-inversion mapping $\Fc(\xv_0)$, but their method requires optimizing the initial noise separately for each prompt. Similarly, \citet{zhou2024golden} aims to train a neural network to refine initial noise, yet this approach depends on a large paired noise and prompt dataset and supervised learning. By contrast, our method needs only standard prompt datasets, which are far more readily available, and does not require specialized noise and prompt annotations.

\section{Additional Experiment}\label{sec:more_results}

\begin{table}
    \centering
    \caption{Training hyperparameters}
    \begin{tabular}{l|l}
     \hline 
         Parameter &  Value \\ \hline
         gradient clip norm & 1.0 \\
         KL regularizer weight, $\gamma_1$ & 1.0 \\
         Entropy regularizer weight, $\gamma_2$ & 0.1 \\
         PPO epochs, $K$ & 4 \\ \hline
    \end{tabular}
    \label{tab:training_hyperparameters}
\end{table}

In this section, we present additional experimental results for SDXL-Turbo~\citep{sauer2024adversarial} to further validate the effectiveness and robustness of Noise PPO. By default, we set the number of inference steps to 4. We conduct both quantitative and qualitative comparisons on the PartiPrompts and HPSv2 benchmarks, following the same evaluation protocols as described in the main text.

\subsection{Quantitative Comparison}
We report the quantitative results of Noise PPO for SDXL Turbo on the PartiPrompts and HPSv2 benchmarks in Tables~\ref{tab:parti_turbo_results} and~\ref{tab:hpsv2_turbo_results}, respectively. For each benchmark, we evaluate both zero initialized and non-zero initialized policies and compare them to the standard Gaussian baseline. Across all metrics, including Aesthetic Score, PickScore, and HPSv2, Noise PPO with zero initialization consistently outperforms the baseline. Additionally, the non-zero initialized variant improves performance on aesthetic scores and on the concept-art and painting categories of the HPSv2 benchmark, further demonstrating the generality and robustness of our approach.

\subsection{Qualitative Comparison}
\begin{figure}
    \centering
    \includegraphics[width=0.58\linewidth]{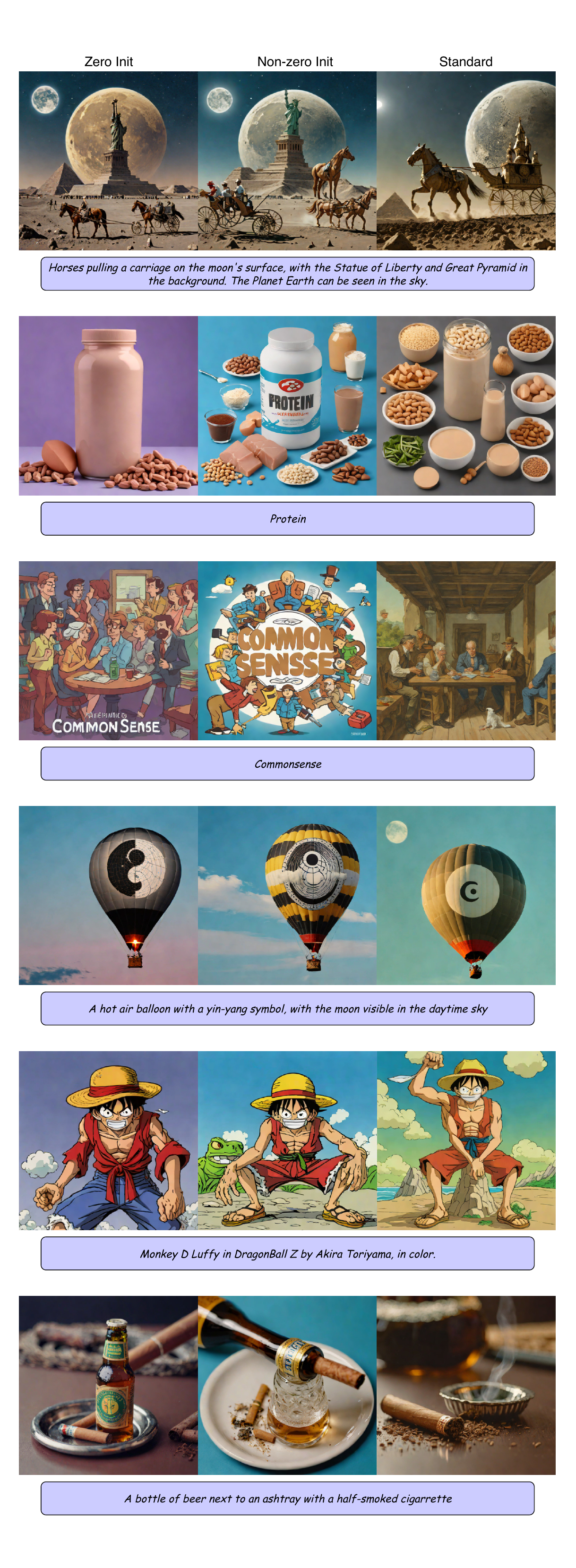}
    \caption{Qualitative comparison. ``Zero Init'' denotes experiments with zero initialization of the final output layer, while ``Non-zero Init'' refers to randomly initialized output layers. ``Standard'' corresponds to images generated from standard Gaussian noise.}
    \label{fig:turbo_comparsion}
\end{figure}

To complement the quantitative results, we provide additional qualitative comparisons in Figure~\ref{fig:turbo_comparsion}. These figures showcase representative samples generated by the baseline model and by Noise PPO with different initialization strategies, using the same random seeds for fair comparison. The visual results demonstrate that Noise PPO produces images with improved text-image alignment and aesthetic quality, and is effective in reducing artifacts and failure cases observed in the baseline generations (e.g., the ``three-armed Luffy'' example). These qualitative findings are consistent with our quantitative analysis and further highlight the practical benefits of optimizing the initial noise distribution.

\begin{table}
    \centering
    \caption{Comparison results on PartiPrompts, including Aesthetic Score, PickScore, and HPSv2. Results are measured over 10 different seeds, with standard deviation (std) included.}
    \begin{tabular}{l|l l l}
        \hline
        Method & Aesthetic & PickScore & HPSv2 \\ \hline
        SDXL Turbo &$5.704 \pm 0.004$ & $22.574 \pm 0.013$ & $28.779 \pm 0.015$ \\ \hline
        Noise PPO (non-zero init) & $\mathbf{5.774 \pm 0.007} \uparrow$ & $22.328 \pm 0.007$ & $28.476 \pm 0.016$ \\
        Noise PPO (zero init) & $5.772 \pm 0.003 \uparrow$ & $\mathbf{22.600 \pm 0008} \uparrow$ & $\mathbf{28.819 \pm 0.013} \uparrow$ \\ \hline
    \end{tabular}
    \label{tab:parti_turbo_results}
\end{table}

\begin{table}
    \centering
    \caption{Comparison results on the HPSv2 benchmark. For fairness, methods fine-tuned solely on the HPSv2 score are excluded. Results are averaged over 10 seeds, with standard deviation (std) reported.}
    \begin{tabular}{l|l l l l}
        \hline
        Method & Anime & Concept Art & Painting & Photo \\ \hline
        SDXL-Turbo & $29.387 \pm 0.018$ & $28.513 \pm 0.013$ & $28.680 \pm 0.023$ & $28.919 \pm 0.026$ \\ \hline
        Noise PPO (non-zero init) & $29.179 \pm 0.005$ & $28.530 \pm 0.018 \uparrow$ & $28.763 \pm 0.011 \uparrow$ & $28.746 \pm 0.013$ \\
        Noise PPO (zero init) & $\mathbf{29.441 \pm 0.007} \uparrow$ & $\mathbf{28.642 \pm 0.017} \uparrow$ & $\mathbf{28.817 \pm 0.008} \uparrow$ & $\mathbf{29.100 \pm 0.015} \uparrow$ \\
        \hline
    \end{tabular}
    \label{tab:hpsv2_turbo_results}
\end{table}

\section{Future Work}

There are several promising directions for future research. First, it would be valuable to explore more expressive policy architectures or alternative RL algorithms to further enhance the effectiveness of initial noise optimization. In particular, investigating the use of GRPO for diffusion model fine-tuning is an interesting avenue, as GRPO does not require training an additional value network. However, determining an appropriate group size in GRPO presents its own challenges: too small a group size may introduce bias in value estimation, while too large a group size can lead to significant GPU memory consumption. We leave a comprehensive study of GRPO-based fine-tuning for future work.

Second, investigating the applicability of the golden noise hypothesis in other generative domains, such as audio or video diffusion models, could broaden the impact of this approach. Additionally, studying the interplay between noise initialization, reward design, and model architecture may yield deeper theoretical understanding and practical guidelines for RL-based fine-tuning. We hope this study will inspire continued research into minimalist RL strategies and the fundamental role of noise in generative modeling.

\section{Limitations}\label{sec:limitations}
While Noise PPO offers a minimalist and effective approach to fine-tuning diffusion models, our study has several limitations. First, the improvements from optimizing the initial noise distribution are most pronounced at low inference steps, and the benefits diminish as the number of inference steps increases. This suggests that the influence of the initial noise is gradually washed out by the iterative denoising process, limiting the practical impact of noise optimization in high-step regimes (Figure ~\ref{fig:diff_infer_parti} and ~\ref{fig:diff_infer_hpsv2}).

Second, although non-zero initialization can yield persistent gains for certain metrics such as aesthetic quality, the choice of initialization and its interaction with different reward functions are not yet fully understood. Third, our experiments focus primarily on text-to-image diffusion models and a specific set of reward functions; it remains to be seen whether the golden noise hypothesis and the effectiveness of Noise PPO generalize to other modalities, tasks, or more diverse reward designs.

Finally, we model the noise policy as a Gaussian distribution for simplicity, which may restrict the expressiveness of the learned noise generator. Exploring more flexible policy architectures or alternative optimization strategies could further enhance performance.

\section{Broader Impacts}\label{sec:broader_impacts}

Our primary aim is to advance the fundamental understanding of generative models, and we believe our findings will be beneficial to the research community. An immediate application of our method is its extension to large-scale visual generation models, such as text-to-image or text-to-video diffusion models. By providing a minimalist and efficient RL-based fine-tuning approach, our work has the potential to reduce the computational cost associated with training and inference in these models. Furthermore, our results may inspire new directions in reward-driven generative modeling and the design of more controllable generation systems.
On the negative side, as with all generative models, our method learns statistics from the training data and may therefore reflect or amplify biases present in the data. Additionally, improved generative capabilities could be misused to create misleading or harmful content, such as disinformation or deepfakes. We encourage practitioners to consider these risks and to use our methods responsibly, with attention to ethical guidelines and potential societal consequences.